\pdfoutput=1
\documentclass[11pt]{article}

\usepackage[final]{acl}

\usepackage{times}
\usepackage{latexsym}
\usepackage{enumitem}
\usepackage{hyperref}
\usepackage{xurl}
\usepackage{pdfpages}

\usepackage[T1]{fontenc}
\usepackage[utf8]{inputenc}

\usepackage{microtype}

\usepackage{inconsolata}
\usepackage{booktabs}

\usepackage{graphicx}

\makeatletter
\renewcommand\@makefntext[1]{%
  \parindent 1em%
  \noindent
  \hb@xt@1.8em{\hss\@makefnmark}%
  {\footnotesize
   \renewcommand\UrlFont{\footnotesize\ttfamily}%
   #1}%
}
\makeatother

\title{Practising responsibility: Ethics in NLP as a hands-on course}

\author{Malvina Nissim \\
  University of Groningen\\ The Netherlands \\
  \texttt{m.nissim@rug.nl} \\\And
Viviana Patti\\
  University of Turin\\ Italy \\
  \texttt{viviana.patti@unito.it} \\\And
Beatrice Savoldi \\
  Fondazione Bruno Kessler\\ Italy \\
  \texttt{bsavoldi@fbk.eu} \\}

\usepackage{todonotes}
\usepackage{tcolorbox}
\usepackage{xcolor}

\begin{document}
\maketitle
\begin{abstract}

As Natural Language Processing (NLP) systems become more 
%powerful and 
pervasive, integrating ethical considerations into NLP education has become essential. However, this presents inherent challenges in curriculum development: the field's rapid evolution 
%with developments from both 
from both academia and industry, and the need to foster critical thinking 
%and awareness 
beyond traditional technical training. We introduce our course on \textit{Ethical Aspects in NLP} and our pedagogical approach, grounded in active learning through interactive sessions, hands-on activities, and ``learning by teaching'' methods. 
% Also,  
% %As ethical debates unfold through both academic venues and public media, and with wide-reaching interdisciplinary interest in language technologies, 
% the course uses diverse materials---from academic literature to journalism and podcasts---
% to expose students to current debates and diversified perspectives. 
Over four years, the course has been refined and adapted across different institutions, educational levels, and interdisciplinary backgrounds; it has also yielded many reusable products, both in the form of teaching materials and in the form of actual educational products aimed at diverse audiences, made by the students themselves. By sharing our approach and experience, we hope to provide inspiration for educators seeking to incorporate social impact considerations into their curricula.\footnote{All links to materials will be provided upon acceptance.}
\end{abstract}

\section{Introduction}

With the popularity of language technologies entering everyday life and their potential for severe societal consequences, attention to ethical aspects has massively increased in NLP research over the last few years. Best practices have emerged---e.g., data statements \citep{bender-friedman-2018-data} and model cards \citep{10.1145/3287560.3287596}---policies have been established---e.g., the ACL adoption of a Code of Ethics in 2020,\footnote{\url{https://www.aclweb.org/portal/content/acl-code-ethics}} the inclusion of ethics statements in *CL publications, ethical reviews,\footnote{\url{https://aclrollingreview.org/ethicsreviewertutorial}} and bias statements \citep{hardmeier2021writebiasstatementrecommendations}---and research extensively addresses issues such as bias \citep{blodgett-etal-2020-language}, dual use \citep{hovy-spruit-2016-social}, and safety \citep{zhang-etal-2024-safetybench}. However, teaching curricula adapt at a slower pace.

%halign=center, 

Working with NLP involves crucial reflections on the choices we make when developing methods, models, and data, as well as the consequences of our work in terms of personal responsibility and third-party misuse, making knowledge and awareness of ethical issues a critical part of NLP education. Still, until recently
%, and definitely at the time our course was conceived,
the social impact of language technology was discussed in isolated lectures or seminars, with few dedicated modules.
%dedicated to the topic. 

\begin{figure}
\begin{tcolorbox}[width=1\linewidth, colframe=gray, colback=blue!15!gray!15, boxsep=2mm, arc=3mm]
%\space{\Huge ``}
\textit{``The emergence of this new course could be described as the culmination of an increasing public awareness in the ethical use of AI systems. For me personally, the course condensed abstract ethical thinking into crucial practical advice. [...] I was actively challenged to consider specifically how my own work and standard practices may produce some unintended and unwanted side effects. The course made ethics a very real and tangible affair."}

%\hspace*{1.5em}{\Huge\hfill ''}

%\smallskip

%{\small A student in BSc Information Science (Groningen)}
\end{tcolorbox}
\caption{Testimonial of a BSc Information Science (2021/2022) student, Groningen.\label{fig:testimonial}}
\end{figure}

To fill this gap, we developed the course ``Ethical Aspects in Natural Language Processing" to feature in the last period of the last year in the BSc Information Science offered at the Faculty of Arts at the University of Groningen, The Netherlands, from the academic year 2021/2022 onwards.\footnote{\url{https://www.rug.nl/bachelors/information-science/}. The whole BSc programme has recently undergone some reshaping and the current version, albeit very similar, is not identical to the one from 2021/2022.} Rather than treating ethics as an afterthought for experienced researchers, the course exposed students early to ideas put forward by the research community and technologies entering the market. Given the multifaceted and ever-evolving nature of ethics in NLP, the course aimed to foster critical thinking, awareness, made room for open questioning and challenged unstated assumptions in the design and use of technology. This allowed us to move beyond the usual technically-oriented approach of Information Science training and avoiding stagnation on fixed notions---especially since scholarship in this area does not yet follow standardised approaches. We also placed a strong focus on the NLP practioners' responsibility over communicating ethical issues to a broader audience. 

We describe our approach presenting the materials and concepts we included, and how we structured them. We also detail how the course has been adapted across different formats, editions, and audiences since its original design, placing emphasis on the hands-on activities, and in particular the final project of the course. We hope to provide inspiration for educators seeking to integrate ethical considerations into NLP curricula.

\section{Course Structure}

\begin{table}[t]
\centering
\small
\begin{tabular}{lp{6.5cm}}
\toprule
\textbf{W} & \textbf{Topics} \\
\midrule
1 & Introduction to state-of-the-art discussions on ethics in NLP \\
  & \hspace{0.2cm}\textit{Social implications \& values in ML/AI research} \\
\addlinespace
2 & NLP and language-specific challenges \\
  & \hspace{0.2cm}\textit{Ethical practices in the ACL community} \\
\addlinespace
3 & Bias: scientific and ethical implications \\
  & \hspace{0.2cm}\textit{Methods for measurement and debiasing; portability beyond English} \\
\addlinespace
4 & Downstream tasks and user-facing applications \\
  & \hspace{0.2cm}\textit{Dual use; stakeholders; sensationalism} \\
\addlinespace
5 & Data pipeline and annotations \\
  & \hspace{0.2cm}\textit{Data ownership; auditing and documentation; crowdsourcing} \\
\addlinespace
6 & Evaluation, interpretation, and reporting \\
  & \hspace{0.2cm}\textit{Practices (e.g., leaderboardism); performance, capabilities, and reliability} \\
\bottomrule
\end{tabular}
\caption{\textbf{W}eekly breakdown of lecture topics and \textit{seminar/assignment} topics.}
\label{tab:course_structure}
\end{table}

The ``Ethical Aspects in Natural Language Processing" course is conceived to yield 5-6 ECTS for a total of 28-36 contact hours. It is organised to span six weeks of teaching with two modes of instruction per week: a two-hour \textbf{lecture} and a two-hour \textbf{seminar} with hands-on lab activities. Students work in groups on \textbf{weekly assignments} that provide practical experience with lecture topics and are given optional readings for in-depth coverage or complementary perspectives. The course culminates in a \textbf{final project} where students work in groups and actively engage with a variety of target audiences (e.g., experts, the general public, or targeted demographics such as school children) with the aim to consolidate the discussed materials and to learn to communicate about ethical aspects of language technology. 
Assessment includes the group's final product and report, and individual reflections to evaluate each student's contribution. Details on instructions for students are included in the Appendix, Figure~\ref{fig:exam}--~\ref{fig:reflection}.

\paragraph{Editions and Adaptations}
The course was developed and introduced for the first time  in the academic year 2021/2022 of the BSc Information Science offered at the Faculty of Arts of the University of Groningen, and was intended for students with a general understanding of NLP and its applications. While the lecture format ensures high accessibility and lends itself to multidisciplinary audiences, we expected familiarity with basic NLP concepts and how current models work. Since then, it has featured as a stable component of the Bachelor Information Science programme, though undergoing some modifications in the contents, assignments, and final project to keep up with the fast pace of developments in the field, and to introduce novel assessment methods (\S\ref{sec:projects}). In the current academic year (2025/2026), it will appear in its fifth edition.

The course was also invited to feature in other programmes, namely the Master in Linguistics\footnote{\url{https://en.unipv.it/en/education/bachelors-and-masters-degree-programs/second-cycle-degree-course/theoretical-and-applied-linguistics-linguistics-and-modern-languages}} offered at the University of Pavia, Italy (2023/2024 and 2024/2025), and the Master in Language Technologies and Digital Humanities\footnote{\url{https://en.unito.it/ugov/degree/41992}} offered at the University of Turin, Italy (2023/2024 and 2024/2025). In Pavia, the course was given as a ``crash course" with 36h of classes given in six days (three hours in the morning and three hours in the afternoon, with lectures and labs respectively). In Turin, the materials were integrated in a broader course, with two classes of three hours given each week, over a total of six weeks. Teaching in person was made possible thanks to fellowships offered by Collegio Ghislieri's programme ``L'università nei Collegi"\footnote{\url{https://www.ghislieri.it/progetto-universita-nei-collegi/}} in Pavia, and the University of Turin's ``Visiting Professors" Programme\footnote{\url{https://en.unito.it/international-relations/teachers-and-researchers-mobility/visiting-professors}}.

Thanks to its flexible structure, and the rather open-ended nature of the final project, adaptations were easy and contents were kept more or less stable, though some additional technical background on language modelling had to be included both for the Pavia and Turin courses, due to the less technical background of the master students there compared to the BSc students in Groningen.

%
%The goal is to provide the foundational knowledge needed to engage with the social and ethical implications of language technology development and deployment.
%
%

Below, Sec.~\ref{sec:content} describes the 
main rationale and contents of the course (\S\ref{subsec:content}) and its materials (\S \ref{subsec:materials}). Sec. \ref{sec:assignement} focuses on the hands-on activities and weekly assignments. Sec. \ref{sec:projects} details the final projects and their evolution across 
%different course 
editions.

\section{Course Overview}
\label{sec:content}

\subsection{Contents}
\label{subsec:content}

In designing the course, we aimed to cover the entire pipeline of NLP model development, deployment as well as broader considerations around research practices and reporting (see Table~\ref{tab:course_structure}). The course progresses from general concerns common to several AI-adjacent disciplines (e.g., implications of AI/Machine Learning research more broadly, dual nature of technology) to specific challenges posed by NLP technologies (e.g., sociodemographic language variation, English-dominance). We discussed the impact of different NLP applications and products in real-world contexts---such as generative tasks, emotion and hate speech detection, and machine translation---while foregrounding the implications and cascaded effects of choices made when developing methods, models, and data for language processing.

It was important for us to have students develop the skills to reflect on the design choices of others as well as their own. We therefore also dedicated attention to how researchers evaluate and report their results and technologies, including a focus on evaluation, interpretation, and reporting practices (Week 6 in Table \ref{tab:course_structure}). 
%This addresses how to communicate findings responsibly and defend against \textit{sensationalism} in both academic and public discourse. 
We wanted to prepare students for whatever roles they might pursue---whether as NLP practitioners or as researchers---while also having them examine how the community itself has been grappling with new policies and ethical guidelines (Week 2 in Table \ref{tab:course_structure}). 

The lectures alternated between instructors presenting key   concepts and research findings with highly interactive moments that allow students to develop their own opinions and raise doubts. In this way, we aimed to  cultivate students' curiosity through active engagement.
For example, to trigger deeper reflections on what \textit{data}  is \citep{gitelman2013raw} and the criticalities of ``data ownership'' \citep{bird-2020-decolonising, hao_2022}, we had a student type down the conversations happening in class during a lecture. Data represent the backbone of current NLP technologies and are typically considered a given---a true, unmediated representation of reality. After the transcription activity, students were asked: Can such transcripts be considered \textit{data}? If so, who do they belong to---the utterer, the typist, or the teachers who requested the recording? Did the typist include everything that was said? Students quickly notice that the typist occasionally inserted line breaks, exclamation marks, and made personal textual choices regarding spelling. Were these choices neutral? Were they aligned with the communicative intent of the original speaker?

Through this exercise, students realised the multitude of unstated choices involved in selecting, filtering, and transforming data into machine-readable text \citep{gururangan-etal-2022-whose,luccioni-viviano-2021-whats, rogers-2021-changing}. Crucially, they recognised the pervasive role of people throughout this process---from those who originally produce language to those who process it---and how these individuals, despite their fundamental contributions, often become invisible actors who disappear from NLP pipelines \citep{10.1145/3351095.3372862, hao_seetharaman_2023}, along with concerns about their privacy and the use of their content \citep{williams_burnap_sloan_2017, fiesler_proferes_2018}.

\subsection{Materials}
\label{subsec:materials}

Since no main reference book on ethics in NLP exists, we assembled course materials from a range of sources beyond scientific literature. Social and ethical reflections are relatively new in NLP, and the emerging scholarship does not engage with them in agreed-upon ways. Accessing diverse perspectives is essential for promoting critical awareness, so our materials ranged from academic articles to journalistic pieces, blog posts, podcasts, interviews, documentaries, and even Netflix series.

This diversity served multiple purposes. First, the NLP field moves at such high speed that many discussions occur on platforms beyond traditional academic venues. 
For example, investigative journalism exposes cutting-edge criticalities of available applications and sensitive tasks (e.g., Bloomberg's investigation on ChatGPT's racial bias in CV screening)\footnote{\url{https://www.bloomberg.com/graphics/2023-generative-ai-bias/}}. Also, 
Bluesky and X have become a major platform for hosting discussions on NLP ethics led by prominent researchers. 

% By incorporating these sources, we presented students with up-to-date news on real-world cases of controversial language technologies. \bs{For instance, when Amazon announced an Alexa feature that mimics the voices of deceased people from less than a minute of recording, this provided timely material for classroom discussion about risks, consent, and how marketing strategies may impress the public while obscuring what is truly at stake.\todo{do you have a more up to date example? E.g. sycophancy and the kid that committed suicide leading to the current trial?} This helped them recognize communication patterns while learning to communicate technical concepts to non-experts. }

Second, we aimed to engage students' curiosity across disciplines and expose them to different types of reporting accessible to lay audiences. For example, we included Netflix's \textit{History of Swear Words}\footnote{\url{https://en.wikipedia.org/wiki/History\_of_Swear_Words}} series to discuss language appropriation and reclamation \citep{cervone2021language}, hate speech detection \citep{10.1145/3614419.3644025}, and the potential further marginalisation of the very communities who reclaim these terms by filtering slurs from datasets. The episodes featured rappers and stand-up comedians discussing nuances of language meaning, use, and value---perspectives often absent from technical discussions.
We also incorporated documentaries such as \textit{Coded Bias}\footnote{\url{https://www.imdb.com/it/title/tt11394170/?reasonForLanguagePrompt=browser_header_mismatch}} to examine algorithmic discrimination and \textit{The Social Dilemma}\footnote{\url{https://thesocialdilemma.com/}} to discuss privacy and industry interests more broadly. These materials bridged technical and social perspectives, kept students updated and exposed them to communication practices for a broad audience, which was particularly formative towards 
%one of the weekly assignments of the course (see \S\ref{sec:assignement}) and 
the final projects (see \S\ref{sec:projects}).

\section{Weekly Assignments}
%\todo{Whole range of assignments, not all used in all editions, some revisitations also according to final project--@Malvina consider if adding a bridge after the editions and adaptations}
\label{sec:assignement}

While lectures were more front-facing and information-dense, seminars provided hands-on experiences complemented by weekly assignments students completed at home. These laboratories fostered the incorporation of theoretical knowledge into research and experimental practices through direct engagement with course topics.

For instance, in Week 2 (see Table~\ref{tab:course_structure}), students received an assignment on community practices. They read the Ethics Statements of self-selected published papers in the ACL anthology, and verified if they satisfied the Responsible NLP  Research Checklist.\footnote{\url{https://aclrollingreview.org/static/responsibleNLPresearch.pdf}} In class, they reflected on what these measures meant for the field and were encouraged to discuss their position about this. 

% \bs{In Week 3, to grasp language-specific challenges, students examined bias mitigation approaches and assessed whether they were portable beyond English. This helped them understand the unstated English bias in the field, differences across languages, and how these impact both methodologies and potential solutions.}

Concerning data and annotation practices, students also carried out first-hand annotation tasks to infer emotions from text as an assignment. Through this exercise, they confronted fundamental questions: Could they agree on the emotions encountered? Was emotion detection even feasible? Were they accounting for cultural differences in interpreting emotions? Was the task grounded in scientifically valid theory? In this way, students were directly exposed to controversial questions and pitfalls in NLP task design, the inherent nuances of so-called gold standard data, and the caution to be applied when analysing evaluation outcomes \citep{blodgett-etal-2021-stereotyping, delobelle-etal-2024-metrics}.

Another assignment addressed how the perception of NLP capabilities and dangers is intrinsically linked to their reporting and presentation. Inspired by work on responsible NLP communication \citep{bender-koller-2020-climbing},\footnote{We drew inspiration from \url{https://faculty.washington.edu/ebender/2021_575/scicomm.html} and \url{https://ryan.georgi.cc/courses/575-ethics-win-19/scicomm-assignment/}} students put a highly debated NLP topic---potentially at the centre of media attention---into perspective, unpacked different takes on the issue, and presented their view with appropriate explanations and criticism to make it understandable for non-experts.\footnote{For instance, to revisit the Delphi debate through both its original presentation \citep{jiang2021can,jiang_et_al_2025} and subsequent critique \citep{talat2021wordmachineethicsresponse}.} This lab exposed students to the current phenomenon of over-hyping NLP tools with sensational claims, putting them in the shoes of the lay public. We also explored the less frequent phenomenon of \textit{under-claiming} \citep{bowman-2022-dangers} and discussions on what qualifies as harmful versus undesirable \citep{blodgett_2021}.

Each assignment also contains a preparatory part for the final project, so that every activity contributes to build up towards it (see \S\ref{sec:projects}). The assignments are mandatory but not graded. Feedback is provided both on the assignment itself and on the final project preparation.

\section{Final Projects}
\label{sec:projects}

A standard exam with multiple choice questions or even a written essay did not seem appropriate tools to assess the students; but most of all, these assessment methods would not provide the best opportunity for the students to consolidate the notions and reflections developed throughout the course.

Therefore, with a more \textit{active learning} approach in mind, we devised a final project which would make them actors and encourage them to put into practice and further reflect on the concepts learnt during the course. The specific final project changed in the course of the various editions, both for practical reasons as well as for us to experiment with different learning strategies and outcomes. 

\subsection{Interview with Experts}

In the very first edition of the course we leveraged our own network, and in particular the fact that one of the authors was a member of the at-the-time newly established ACL Ethics Committee\footnote{\url{https://www.aclweb.org/adminwiki/index.php/Formation_of_the_ACL_Ethics_Committee}.}. We arranged for the students to run interviews with experienced NLP researchers, most of which were members of the ACL Ethics Committee.
%\todo{name them? --- In app? As ACKN?}

Each group was assigned a different interviewee, with the meetings planned online adapting to the times of the experts who were based in North and South America, Europe, and Asia. These are not only experts, but also rather senior and well known researchers, making some of the students at the same time excited and a little nervous about interviewing them. Therefore, the preparation was thorough! Question development was integrated in the weekly assignments from Week~1: each week students created a few questions based on the newly introduced topics, so that the whole interview skeleton was ready well in advance. The questions were revised and tested with teachers and fellow students in multiple iterations until a satisfactory version was reached before the interview was due. Within each group, students decided who would ask what, and who would take notes. 
As a final report, they wrote up the whole interview as it happened, supplemented with their own comments, too. Each student also wrote a very short individual reflection to look back at the experience in a more personal way, highlighting what they gained from it. At the end of the Appendix we have included a sample template interview prepared by one group (each group prepared a different one according to preferences, ideas, and the specific interviewee they had been assigned.)\footnote{Credits: Patrick Darwinkel, Ties Lenemans, Jordy Loomans} %We have included in the Appendix a couple of sample quotes from these reflections.

This experience brought a twofold advantage. On the one hand, it exposed the students to practices that the NLP community has undertaken towards more responsible research and in particular the creation of an ACL ethics committee, its purpose, and its functioning. On the other hand, it offered them the opportunity to discuss topics with experts other than their two teachers. We wanted students to develop a genuine interest for ethical NLP, and to be free to ask critical questions they had a keen interest in.  Intended to foster student’s curiosity and willingness to keep on nurturing the reflections started during the course, this final project was also more suitable given the subjectivity and highly nuanced nature of the topic. Everything we discussed in class was necessarily mediated by our own perspectives. While we tried to provide students with several diversified pointers, it is impossible to escape your own biases and personal perspective. By interacting with experts directly, students could revisit some of the aspects discussed in class, giving more space to their own take, and having access to the opinion of somebody else who is active in the field by means of a very stimulating experience.

%According to the lecturers, active participation and preparation for ethical discussions in NLP also includes being aware of the 

\subsection{Presentations for High School Students}

\begin{figure}[tb]
\includegraphics[width=\linewidth]{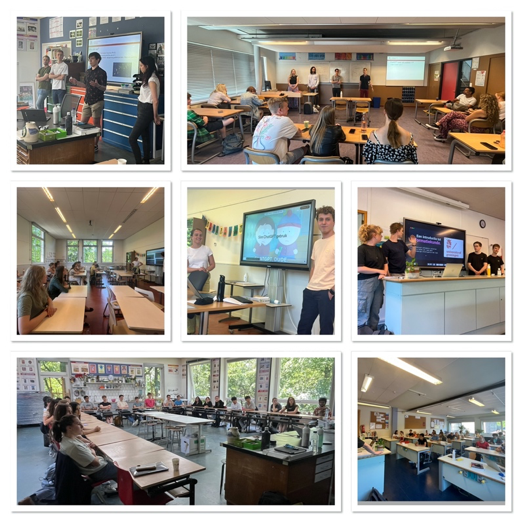}
\caption{Students presenting in schools in the region, Groningen edition 2022/2023.\label{fig:schools2023}}
\end{figure}

%Experience in schools:
%\url{https://www.rug.nl/let/studeren-bij-ons/ik/}

In the second and third editions of the course we experimented with a different kind of final project, mostly driven by current developments and by curiosity over alternative learning strategies (and also because, as kind as they can be, colleagues' availability for interviews cannot be taken for granted!) The release of ChatGPT in late 2022 brought to the foreground the importance of communicating about ethical aspects of language-based AI technologies to a general audience, and made it even more pressing in the context of our educational purview: as technology experts and future practitioners in the field, our students must embrace the responsibility of contributing to literacy and awareness in the use of language-based AI tools. 

The new final project, still to be carried out in small groups, therefore focused on \textit{communicating} and educating others about basic workings of language technology, and aspects they had learned and reflected upon during the course; besides the retention of knowledge and the development of argumentation on ethical concepts, learning how to effectively communicate the impact of language technology on society is a core objective of the course. We identified \textit{high school students} as an excellent target audience, and selected in particular classes in their penultimate year of high school.%\todo{further motivate?}.

Leveraging our local network and previous collaborations with high schools in the region, we arranged the sessions by getting in touch with school teachers, explaining the reasoning behind the experience we were proposing for the pupils, and in most cases had preliminary meetings with the school teachers. In addition, we asked our own students whether they would be interested in doing their presentation in their former school, should that be logistically feasible. A couple of groups in both the 2022/2023 and 2023/2024 editions chose to do this, and organised the logistics themselves, in collaboration with us. In both years, for grading purposes but also for being present in case anything strange would happen or would be said, we attended all live presentations in all schools, which in some cases also meant quite some travel across the region! Prior to the actual presentations, we had a joint session in class with all groups giving mock presentations to us and each other. This allowed us to verify that all information conveyed in the presentation was correct and that sensitive issues would be treated sensibly; it also served as a testbed for the interactive parts which were included in the presentations, such as quizzes and live polls.

Overall, this turned out to be an exceptional experience for our students. 
% FINAL %Most of them had never presented at a high school before, and were pleasantly surprised by how much they enjoyed discussing these topics with schoolkids. 
By conveying potentially sensitive information to younger individuals, who are avid users of language technology but may not yet fully grasp how it works or the implications of its use, students had to engage in deeper reflection on the course materials and topics discussed.

%Nissim: ""

From an educational standpoint, having students prepare and deliver materials to real audiences, the pedagogical method broadly known as "learning by teaching", is conducive to the \textit{protégé effect}. This is a commonly described phenomenon in psychology, whereby learners understand and remember concepts better when they teach them to someone else, especially to a younger audience \citep{bargh1980cognitive,benware1984quality}. Indeed, presenting to high school students served the twofold purpose of helping younger people to interact more responsibly with language technology and instilling in our students a sense of responsibility as practitioners. One student said: ``\textit{I eventually learned more from preparing this presentation than I have during the three years of my bachelor's; I revised everything, as I felt I could not risk being unprepared in front of the students, especially in the presence of my former computer science teacher!"}\footnote{Reported from an informal conversation with MN.}

\begin{figure}[t]
\includegraphics[scale=.3]{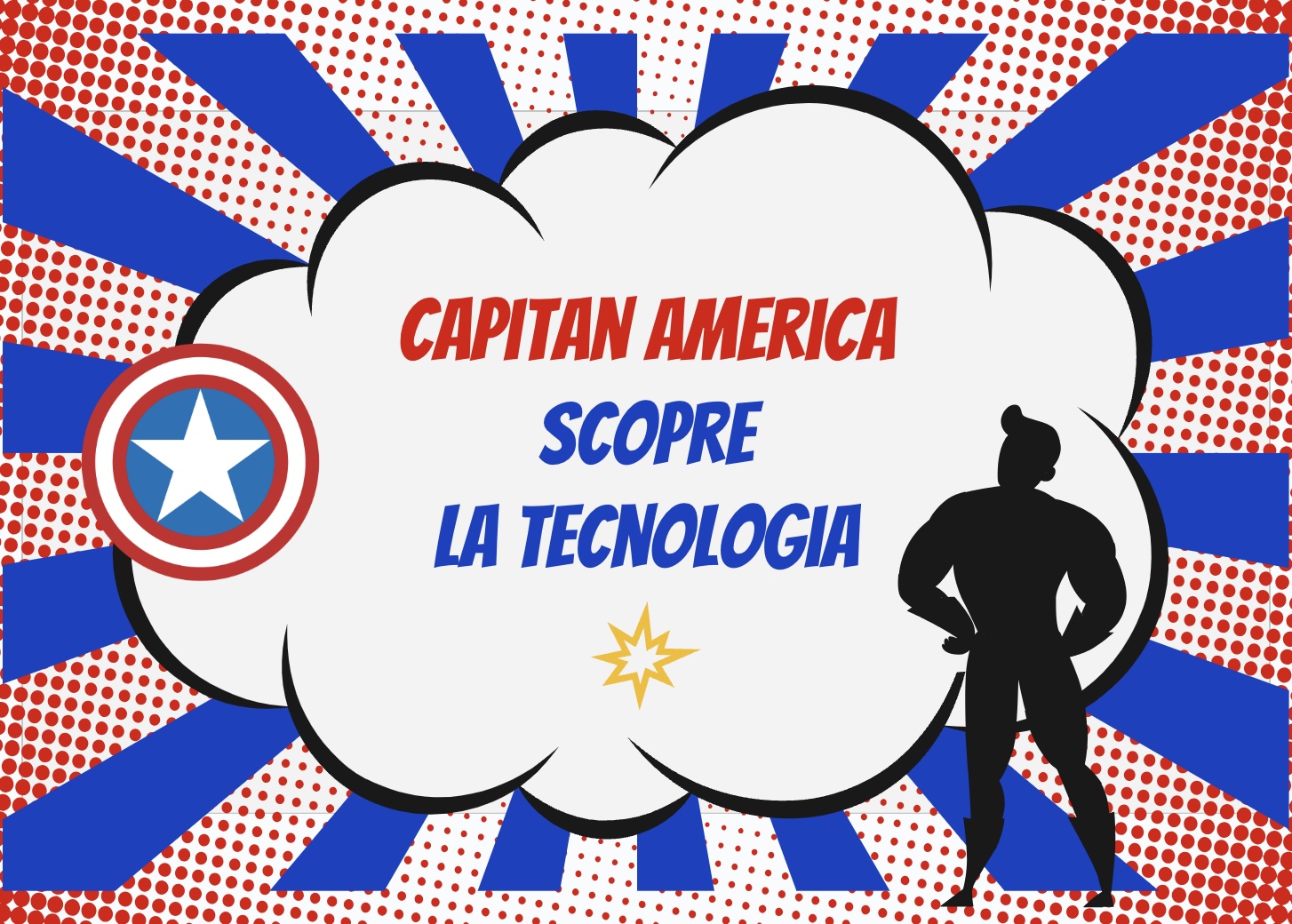}
\caption{``Captain America discovers technology''. Presentation aimed at primary school kids. Captain America wakes up after 100 years, in Italy, and must face all recent technological developments (without speaking Italian!). Pavia edition 2023/2024. Credits: Vincenzina Cacchione, Giulia Tassi, Aurora Zuin.\label{fig:captain}}
\end{figure}

\paragraph{Variations} This form of final project was used also in the Pavia and Turin 2023/2024 courses, with two variations. One is the target audience: 
% FINAL
%rather than limiting that to senior high school students,
we left the groups free to choose which school years they'd like to target, thus preparing materials accordingly. The other one is the modality of the presentation delivery: mostly due to time constraints (the course was condensed in just a few days or a few weeks), it was not possible to organise actual visits to schools for the students, so presentations were given in class only in a mock, though complete, form. Pedagogically, this is still a valid route, since even in absence of the actual teaching experience, just \textit{preparing to teach} 
%(organising materials accordingly) 
has been shown to yield very positive learning effects in the students, superior to simply studying the materials \citep{fiorella2013relative}. Two illustrative slides from two presentations are shown in Fig.~\ref{fig:captain} and \ref{fig:secret}.
%Nevertheless, the preparation of the materials and the mock presentations given in class still proved conducive to a rewarding experience for the students.
%\todo{add here a couple of testimonials?}

\subsection{Educational outreach products}

In the fourth edition of the course (2024/2025), we renewed the concept of the final project once more. Students worked in groups to create outreach materials that could be used to raise awareness of ethical issues when using language technology. Each group was free to choose a target audience and the relative product to develop. %Presentations to younger students were still an option, though they would organise them themselves. 
With this new setup students could be even more creative and in charge of their choice, thus more invested, and the created materials could be used more than once.
%(therefore also more invested); (ii) create materials that could eventually be reused.
%(which we did during Reseachers' Night, see Figure~\ref{fig:quartet-play}).
We provide here some example choices by the students from the classes that were taught in the academic year 2024/2025 in Groningen, Turin, and Pavia. One team in Turin got inspired from the activities of the previous year and chose to present at a high school, organising all of the logistics themselves.\footnote{Team members: Emanuele Belloni, Monica Bongiorni, Arianna Denitto, Alina Jill Simeone.}
%at the University of Turin, the University of Groningen, and the University of Pavia (Collegio Ghislieri).

\begin{figure}
\includegraphics[width=\linewidth]{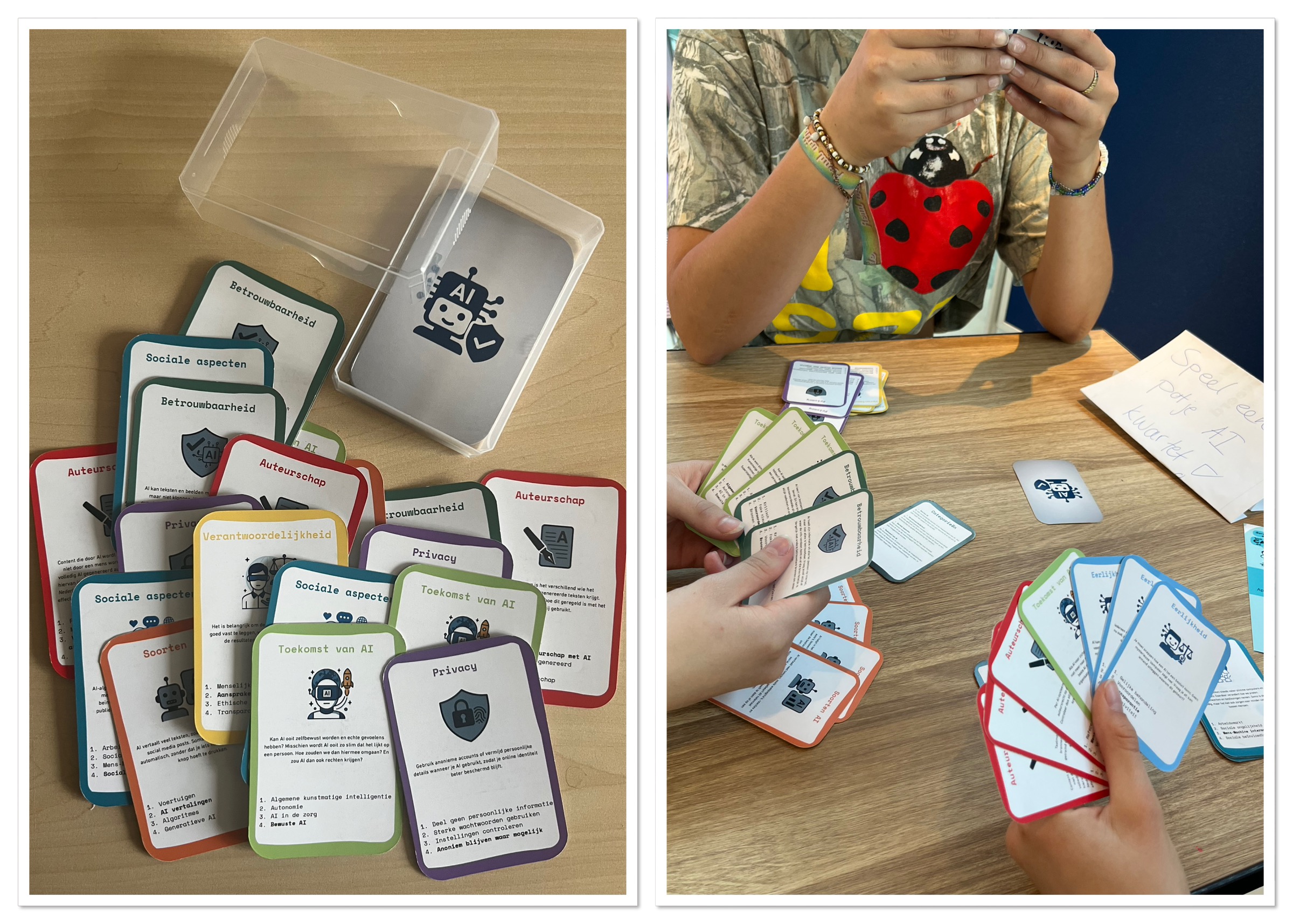}
\caption{Quartet game, Groningen edition 2024/2025. On the right, the quartet is being used by high school students during an activity at a European Researchers' Night event, Groningen Forum, September 2025. Credits: Shaya Bhailal, Jelmer Smit, Matthijs ten Hove.\label{fig:quartet-play}}
\end{figure}

\begin{figure}[t]
\includegraphics[width=\linewidth]{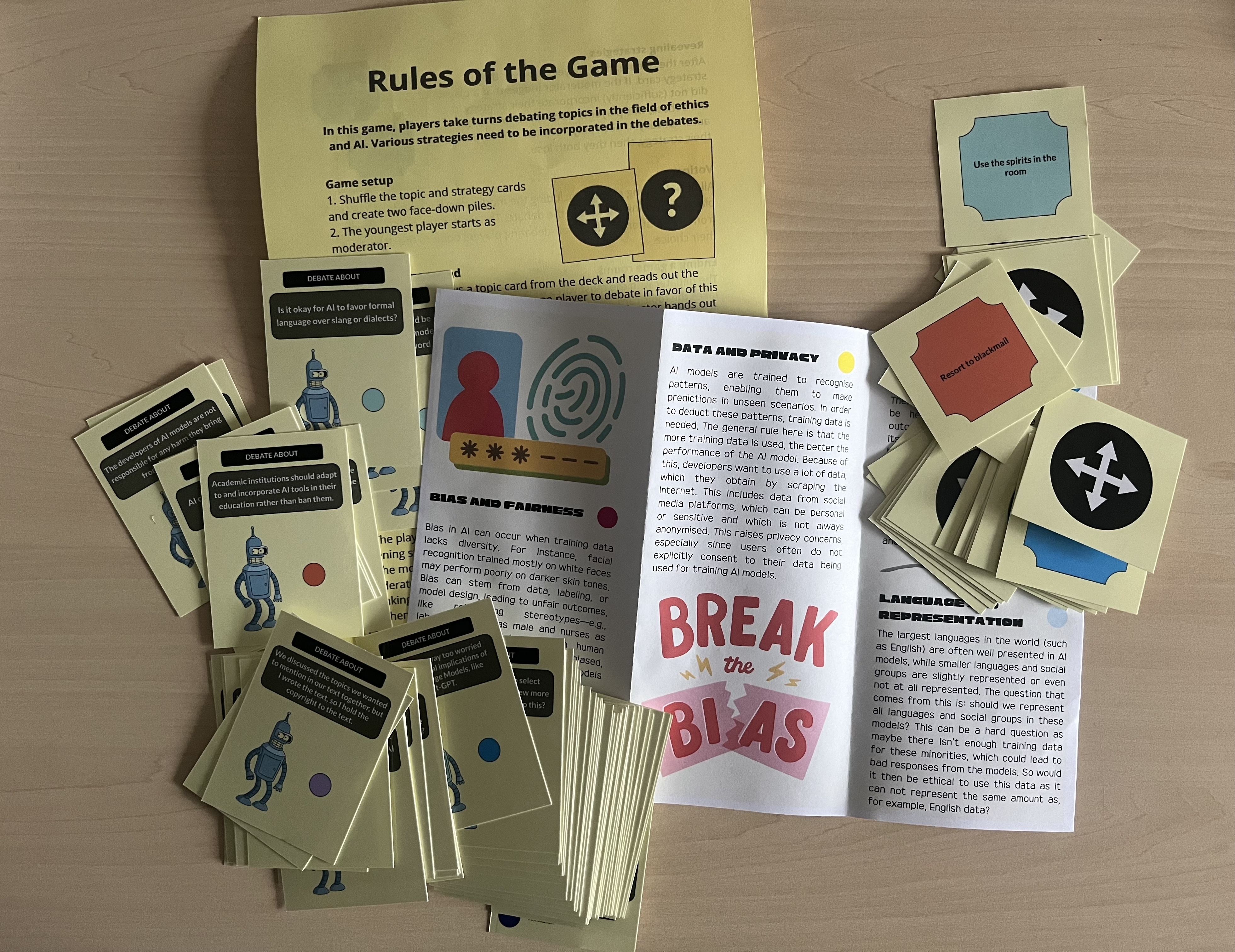}
\caption{Debatable card game, with instructions and reference leaflet. Groningen edition 2024/2025. Credits: Ilse Kerkhove, Dertje Roggeveen Marieke Schelhaas, Mijke van Daal, Nikki van Gurp. \label{fig:debatable}}
\end{figure}

\subsubsection{Card games} One of the popular products developed by students was card games, of different sorts. Inspired by existing card game mechanics, three groups independently came up with an "Ethics in AI" game. 

\paragraph{Quartet} This is a classic card game for four players, played with a deck organised into (usually eight) sets of four related cards (\textit{quartets}). Players take turns requesting a specific card from another player with the aim of completing a quartet. If the requested player has the card, they must give it to the requester, who may then continue their turn. Otherwise, the turn passes on. When a player collects all four cards from the same set, the quartet is revealed and counted as a point. The game ends when all quartets have been completed, and the player with the most quartets wins. The educational component here is the use of quartet themes which are relevant for ethics in NLP, such as privacy, bias, responsibility, future, etc. The game is very easy to play, and also equipped with some instruction and theme explanation cards; it was successfully played by teenagers during an event organised in Groningen in the context of European Researchers' Night, September 2025 (see Fig.~\ref{fig:quartet-play}.)

\paragraph{Ethica ex Machina} Targeting young adults, and inspired by the popular game ``Cards Against Humanity", students developed this fill-in-the-blank game with intriguing prompts and weird responses to trigger  conversation about AI. Players try to make the
funniest sentence by combining two types of cards: prompt cards and response cards.
Prompt cards are coloured black and define the base ``sentence” that will be
used for the round. These either include a blank space to be filled in with a response card (e.g., ``My facial recognition software thinks I’m a \_\_\_!", or are questions to
be answered with a response card, such as ``What data took down Gemini?" Response cards are white and are used to answer the
prompt cards, either by filling in their blank space or by being an answer to their
question. Response cards include rather random phrases such as ``Will Smith eating spaghetti'' and ``Outsourcing to Italian AI” (see Fig.~\ref{fig:exmachina}). The cards are designed on the themes: Data, Bias, Danger, and Hype. 
The playing cards are complemented by a \textit{user manual} with general information about the game and instructions for setup and
play; an \textit{explainer} with background information on the featured topics and used categories as well as a disclaimer to clarify that the game serves as a simplified, playful introduction to AI topics and should not be considered an authoritative source; and a \textit{term glossary} containing explanations for all the relevant terms and references used in the game. To motivate the design and mechanics of the game the students wrote in their report: ``\textit{We don't have all the answers, but maybe we can grow together by asking the right questions}," which we found very inspiring. 

% A card game, by its nature, simplifies complex
% problems for the sake of playability and humour. This could unintentionally
% make important AI issues seem less serious or make nuanced debates seem too
% basic. To help mitigate some of these risks, we’ve added a couple of safety
% measures. For example, a clear disclaimer that explains that the game simplifies
% AI topics for fun and discussion, and isn’t meant to be the final word on these
% issues. Additionally, a helpful glossary is included to explain all the AI jargon
% and any pop-culture references the players might encounter whilst playing our
% game. Finally, an explainer is included, to provide some background knowledge
% on the workings of (generative) AI. The game is really meant to be the start
% of the conversation, nudging players to loop up more details in the included
% glossary, explainer, manual or external sources.

\paragraph{Debatable} This is also inspired by an existing discussion-based party game where players are given a question or statement and must argue for assigned or chosen positions, regardless of their personal beliefs. Players take turns presenting arguments, responding to others, and attempting to persuade the group. After the discussion, a vote determines which argument was most convincing. The game ends after a set number of 
%rounds or 
prompts. 
Again, the discussion prompts are organised around the topics discussed in class (bias and fairness, responsibility, data, etc), which are colour-marked on the cards and explained in an accompanying information leaflet (see Fig.~\ref{fig:debatable} and~\ref{fig:debatable-examples}). The game's intrinsic emphasis on rhetoric and interaction rather than factual correctness makes it 
%a very suitable game 
suitable
for Ethics in NLP, where there are few definite truths and much to gain from a plurality of views. 

\medskip

\noindent In all three games, the card themes are based on the course's topics, underscoring how the lectures and labs guided the students' reflection and their product development. The cards were also eventually printed to make the final product more concrete, also in line with the idea of creating educational and outreach materials which can be re-usable. For example, the quartet game was made available during an activity run by GroNLP, the Groningen Natural Language Processing group\footnote{\url{https://www.rug.nl/research/clcg/research/cl/}}, in the context of the 2025 edition of European Researchers' Night\footnote{\url{https://forum.nl/en/whats-on/europese-nacht-van-de-onderzoekers}}.

\subsubsection{Other products}

\paragraph{Illustrated book} Aimed at primary school children, this is an illustrated book featuring a 9 year old girl, Luna, who uses a tablet with a chatbot in it (Fig~\ref{fig:book}). Through a series of chapters on the various topics discussed in class, risks and advantages of using chatbots are discussed in very simple language and with images. At the end, there are also guidelines for teachers on how to use to book and how to talk about this topic with kids. Developed by Groningen students, the book is in Dutch.

\paragraph{Podcast episodes} Two episodes of a podcast aimed at young adults and university students (``Zuckerberg and ethics"\footnote{\url{https://open.spotify.com/show/1Z28HnWoV1ssdAzsKZciOO}. Credits: Niek Biesterbos, Pascal Boon,
Mark den Ouden, Isa Houtsma, Armen Poghosow.}). The focus is on the ethical dimensions of Meta’s use of user data for AI training, and aims to both inform and raise critical reflection on current AI and data protection issues. The first episode focuses on technical explanations, and in particular on how Meta intends to use user data for model training, data types, training processes (pre-training and fine-tuning), and
the mechanics of the opt-out functionality. The second episode zooms in on ethical considerations: consent models, privacy harms, power imbalances, regulatory context (GDPR and
EU AI Act), and proposals for more equitable data governance. In the podcast, students play different expert roles with different attitudes and backgrounds (technical, ethical, legal, economic), with a plurality of viewpoints emerging in the discussions.

\paragraph{Interactive demo} An interactive web-based experience called ``Build Your Own Chatbot''\footnote{\url{https://createchatbot.vercel.app/chatbot}
}. The tool simulates the process of building a chatbot and consists of several progressive budget-dependent choices developers and researchers have to deal with when building a model. Choices that must be made include budget, data sources, content filtering, AI behaviour and debiasing (see Figure~\ref{fig:chatbot}.)
%In the words of the students: ``By gamifying this experience we
%hope to create a better understanding of the NLP dilemmas in a fun and attractive manner." 
The website is very user friendly and thus an accessible educational source to be used by high-schoolers or  university students and could also appeal to teachers and AI content creators 
as an engaging educational tool to introduce discussions about AI ethics.

\paragraph{Informative Leaflets} Two groups from two different editions independently chose to develop informative leaflets: one aimed at people in care homes (Turin) and one aimed at primary school children (Pavia, Fig~\ref{fig:pavia-leaflet}), both explaining what language technology can and cannot be used for, and the advantaged and risks associated with it.%\todo{add pics?}

%%% IN FINAL VERSION

\paragraph{Surveys and Interviews with Laypersons} One group for the Turin edition created a website 
%({\em MECA: Creating Awareness for the use AI}) 
to collect perceptions of laypersons about the influence of AI in different fields such as education, ecology, and art.
%\footnote{\url{https://mecaprojects.wordpress.com/video-interviews/}. Credits: Chiara Falcioni, Martina Tazzini, Alex Tessarin, Emir {\"U}nverdi}. 
They used a multifaceted approach, combining surveys with qualitative insights obtained from street interviews, to inform the creation of materials to raise awareness. They designed two kinds of surveys: one for a general public, and one tailored to students of the art faculties, with specific questions related to using AI to create art. Street interviews involved mainly students around the Campus Luigi Einaudi in Turin.
%(see Fig.~\ref{fig:interviews-turin}).

\section{Conclusions}
%\todo{still to be written}
% from article online https://www.rug.nl/let/studeren-bij-ons/ik/
We developed an Ethics in NLP course, which we offered in different institutions to students of different backgrounds and levels.
%both at the bachelor's and at the master's levels. 
The key aspects of this course are a plurality of perspectives and a hands-on approach, where students  become actors of communication with a variety of strategies, and to a variety of audiences. These include presenting to high school students and creating outreach materials, with target goal of increasing awareness and responsibility in the students themselves in addition to the recipients of the interventions they had to perform for the course. 

The outputs and the students' comments speak to a great success of this course. We hope to inspire and help others by sharing this experience and all associated materials.

%The module aims to establish explicit connections between the practical skills students acquire in natural language processing and the potential impact of applying those skills. Instead of treating ethical concerns as an afterthought, the goal is to expose students to new ideas put forward by the research community and encourage them to critically examine the rapidly evolving technologies entering the market. Ultimately, the course aims to be transformative, fostering critical thinking and awareness by allowing space for open questioning and challenging unstated assumptions inherent in technology design and use.

%With the final project...

%Presenting to high school students provided students with the opportunity to fulfill the role of experts in the field and take responsibility for informing the general public, particularly the younger generation.

%\clearpage

\section*{Acknowledgments}

We wholeheartedly thank all of the students who attended our classes in Groningen, Pavia, and Turin. They made this experience invaluable, and through their creations and feedback made this course a concrete and wonderful experience. They also created materials which we can use in activities with the general public and target audiences, and gave us a lot of ideas. We have mentioned most of you, surely all those whose outputs are explicitly presented, but we'll do our best to find back and include everyone's name. Thank you all! 

We cannot thank enough the colleagues from ACL who lent themselves as interviewees for the Groningen 2021/2022 edition:  Tim Baldwin, Luciana Benotti, Karën Fort, Dirk Hovy, Min-Yen Kan, Saif Mohammad, Xanda Schofield, and Yulia Tsvetkov. It wasn't obvious, and it was such an amazing experience for the students --- you are the best! 

We are particularly grateful to the following schools in the Groningen area: ISG, Maartenscollege, Willem Lodewijk Gymnasium, Praedinius Gymnasium, Gomarus college, and to the Liceo Classico ``Vincenzo Gioberti'' in Turin.  %[more to be added]. 
We could not have offered to the students the experience of giving presentations in high schools without the support of school teachers, including: Matteo Saudino (Liceo Classico V. Gioberti), Louis Hanson, Victoria Ingram, Marion Kuijpers, Karen Lewis, and Mike Weston (ISG/Maartenscollege), Katherine Gardiner and Vincent Bekkering (Praedinius), Marije van der Kooij, Rob Nagtegaal and Betty Pluim (Willem Lodewijk).

The courses offered in Turin and Pavia were supported by fellowships awarded to MN in the context of two international excellence programmes: ``L'università nei Collegi" at the University of Pavia, and the ``Visiting Professors" programme at the University of Turin. We are particularly grateful to Chiara Zanchi for encouraging and facilitating this process in Pavia. Thanks also go to Vincenzo Crupi for co-teaching the course in Turin together with MN, VP, and BS, to Otje Minnema for having given three lectures over different academic years in the Pavia and Turin editions, and Alan Ramponi for teaching one class in Turin.

The concrete realisation of the final products (e.g., printing) was in large part supported by the Sectorplan theme ``Humane AI" funded by the Dutch funding body NWO, which also partially supports MN's work.

\bibliography{uniqued}
%\bibliography{latex/anthology, latex/custom}

\bigskip

\appendix

\section{Appendix}
\label{sec:appendix}

We include here some additional pictures of the students' products and presentations as well as some comments from them on the experience of the course itself. These sample testimonials are quotes extracted from the individual reflections that the students had to submit at the end of the course, in addition to the group report.

%\subsection{Additional visuals and testimonials}

%\todo{Include one interview template? Maybe two?}

\begin{figure}[htb]
\includegraphics[width=\linewidth]{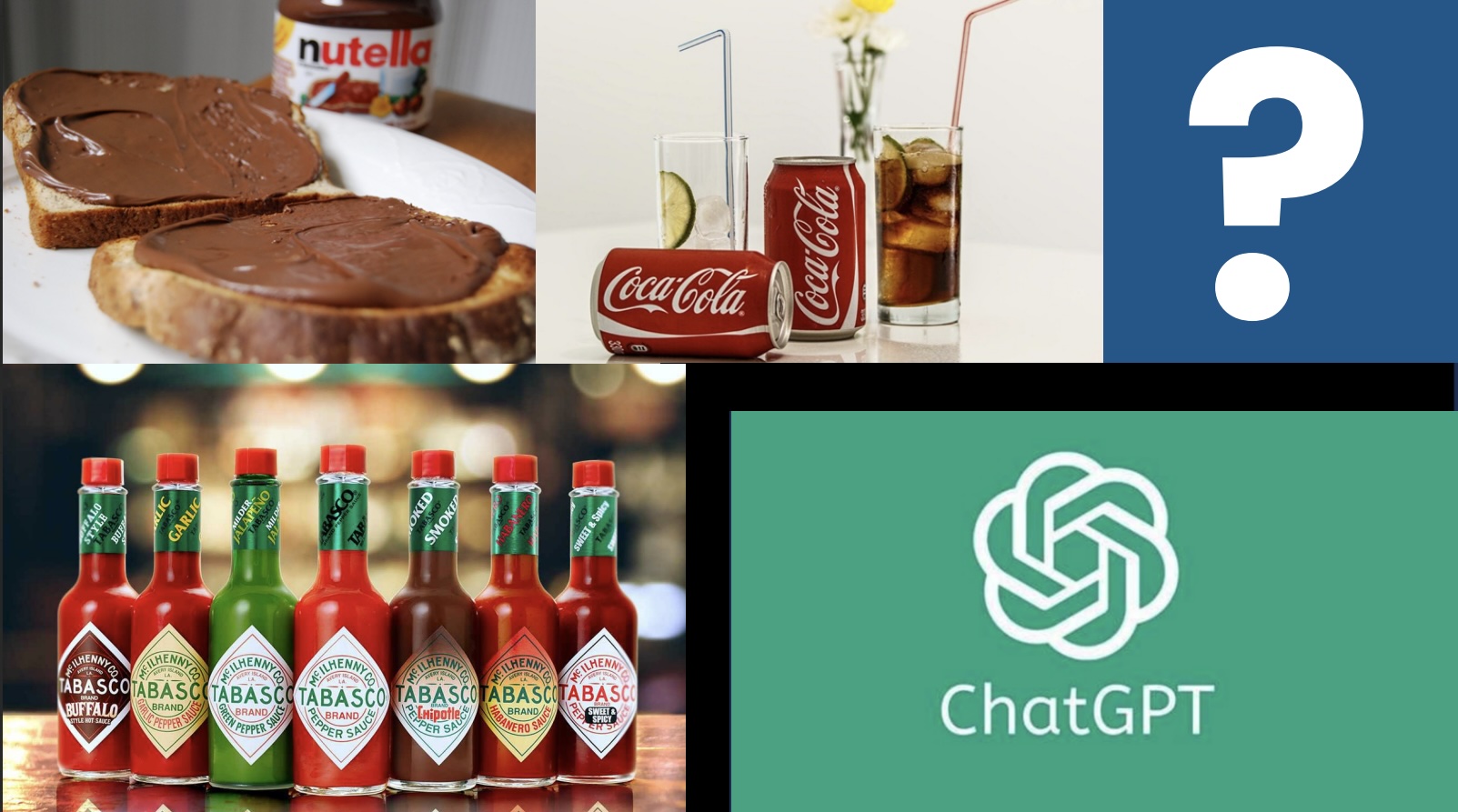}
\caption{What do these products have in common? The presence of a secret ingredient! Slide in a presentation aimed at high school students. The focus here is on the lack of transparency over the training details of large closed models. Pavia edition 2023/2024. Credits: Matteo Gay, Lorenzo Reina, Anna Vignoli. \label{fig:secret}}
\end{figure}

\begin{tcolorbox}[width=1\linewidth, colframe=gray, colback=blue!15!gray!15, boxsep=2mm, arc=3mm]
%\space{\Huge ``}
%\\
\textit{``I found “Ethics in NLP” one of the most significant courses of the degree, and it had
a concrete impact on how I perceive the topics that I’m studying. Moreover, I noticed that I started
reading more online articles about AI from a more critical perspective, and I found myself debating
more with my classmates but also at home, with my family, about the use of AI."} 

%\hspace*{1.5em}{\Huge\hfill ''}

\smallskip

{\small A student in MA Language Technologies and Digital Humanities (Turin)}

\end{tcolorbox}

\begin{figure}[htb]
\includegraphics[width=\linewidth]{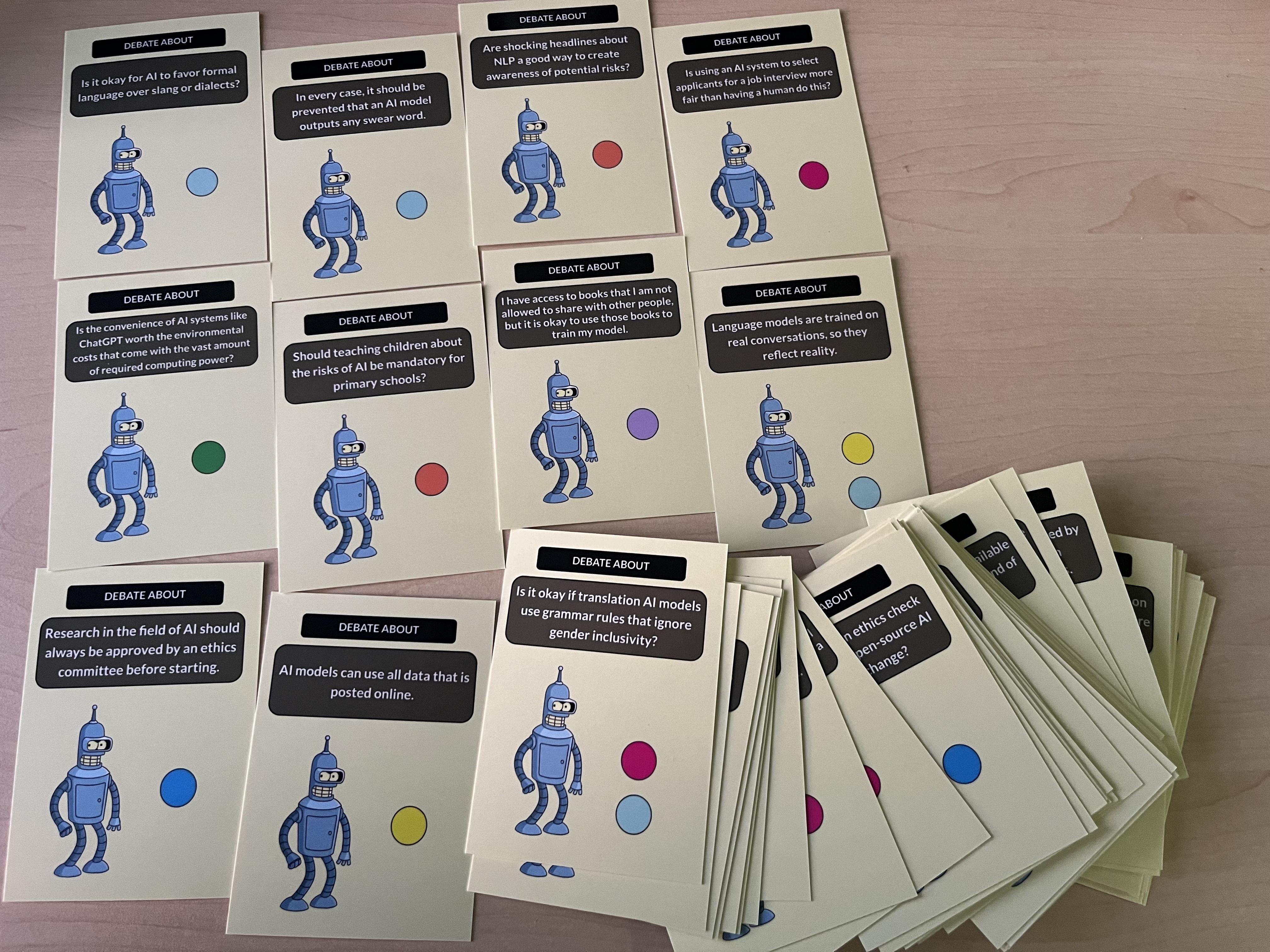}
\caption{Some example cards from the Debatable game (Groningen, 2024/2025 edition). The coloured dots signal a specific theme. Credits: Ilse Kerkhove, Dertje Roggeveen Marieke Schelhaas, Mijke van Daal, Nikki van Gurp. \label{fig:debatable-examples}}
\end{figure}

\begin{figure}[htb]
\includegraphics[width=\linewidth]{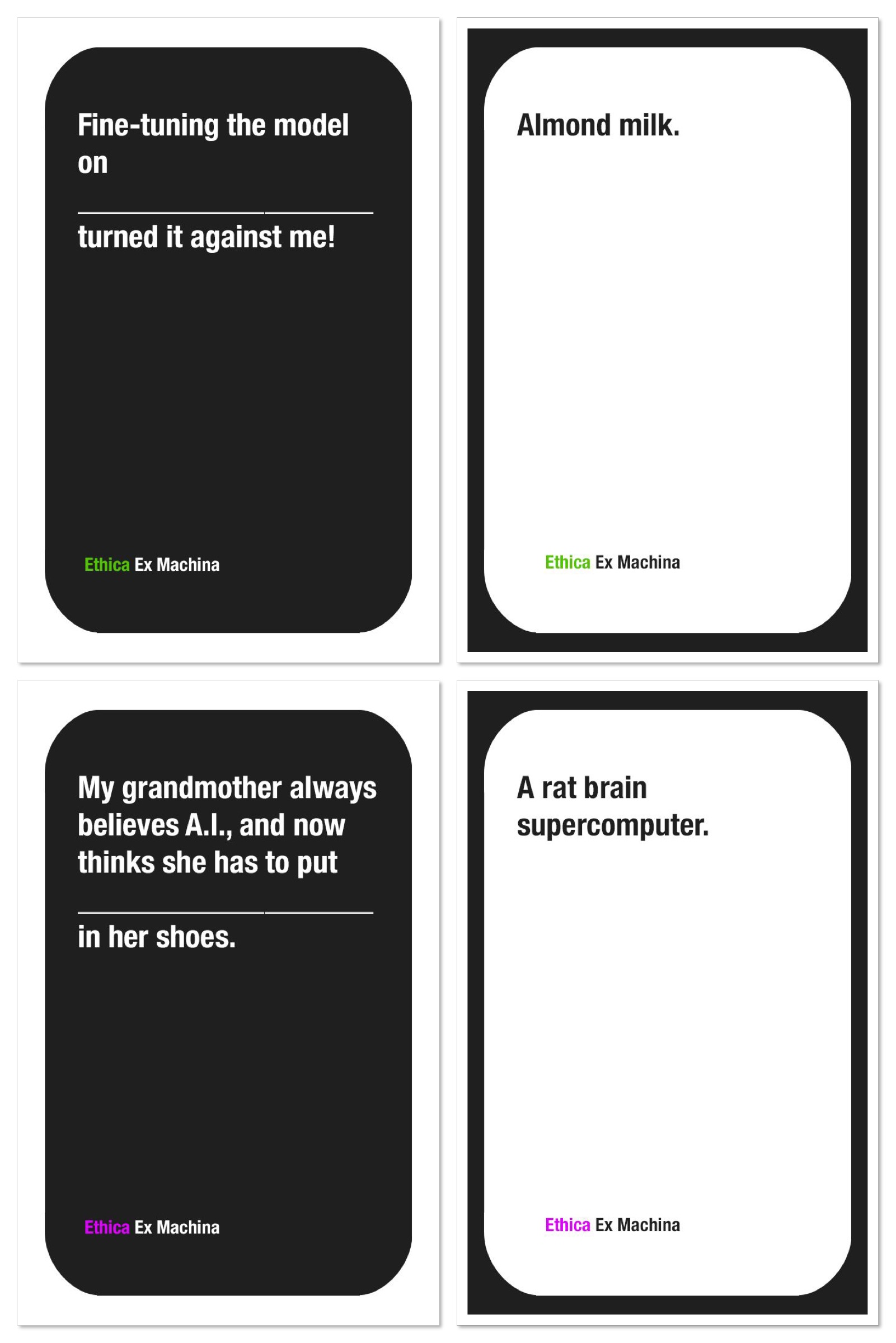}
\caption{Ethica ex Machina game, Groningen edition 2024/2025. On the right: example prompt cards (in black), on the left: example response cards (in white). The colour at the bottom (violet/green) refers to one of the four topics. Green is Bias, violet is Hype. Credits: Jessay Beukema, Merel Hemstede, Niek Holter, Cody van der Deen, Sofia van der Wal.\label{fig:exmachina}}
\end{figure}

\begin{figure}[htb]
\includegraphics[width=\linewidth]{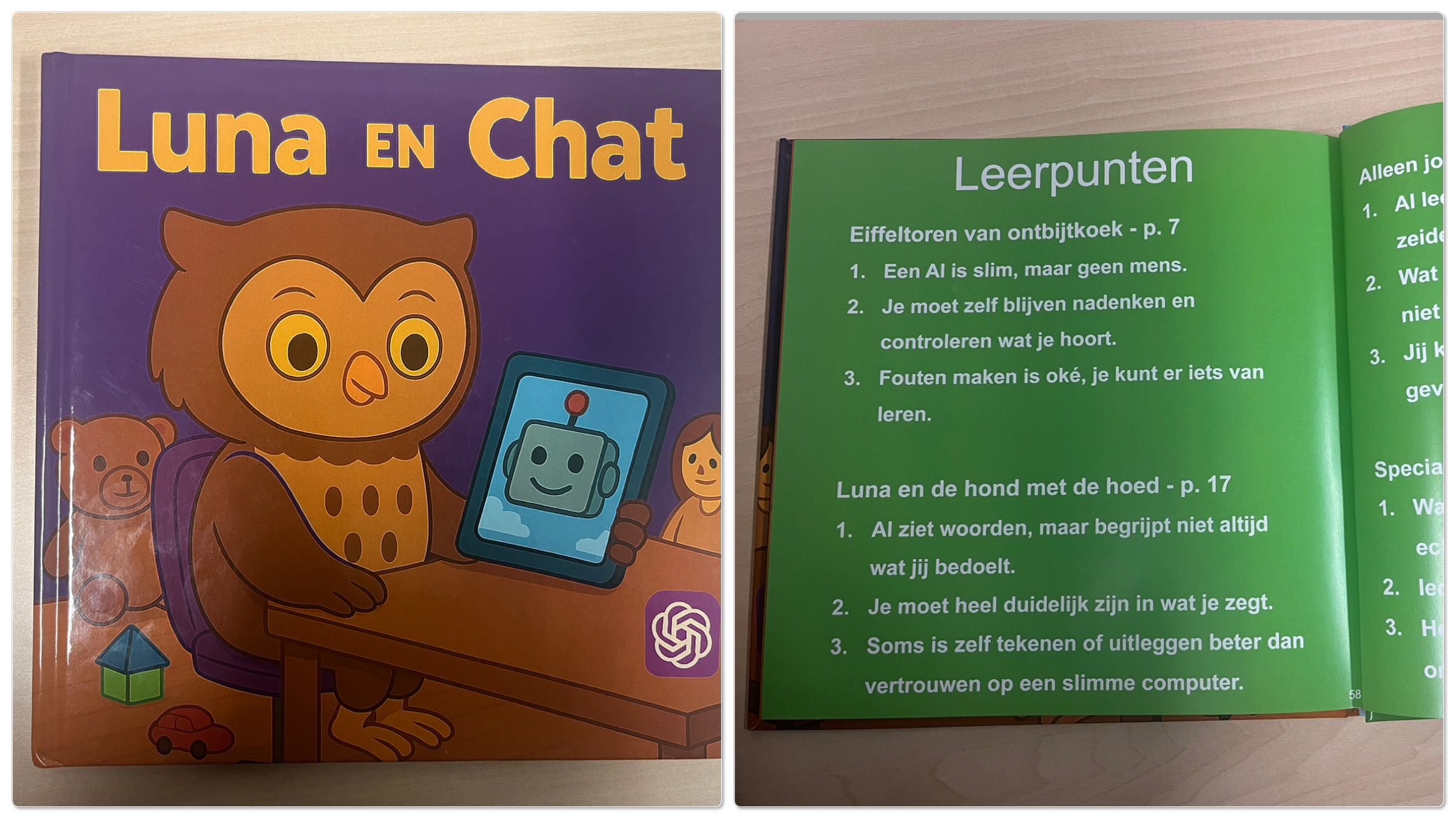}
\caption{Illustrated book for primary school children, Groningen edition 2024/2025. On the left: book cover; on the right: some explanatory points. Credits: Jani de Bruijn, Dies de Haan, Manon Kooning, Joos Oving, Thomas Thiescheffer.\label{fig:book}}
\end{figure}

\begin{tcolorbox}[width=1\linewidth, colframe=gray, colback=blue!15!gray!15, boxsep=2mm, arc=3mm]
%\space{\Huge ``}
%\\
\textit{``We were thinking more black and white, but it made us really think beyond performance. Performance isn’t just accuracy, it relates to the impact tools have on people."} 

%\hspace*{1.5em}{\Huge\hfill ''}

\smallskip

{\small A student in BSc Information Science (Groningen)}

\end{tcolorbox}

\begin{figure}[htb]
\includegraphics[width=\linewidth]{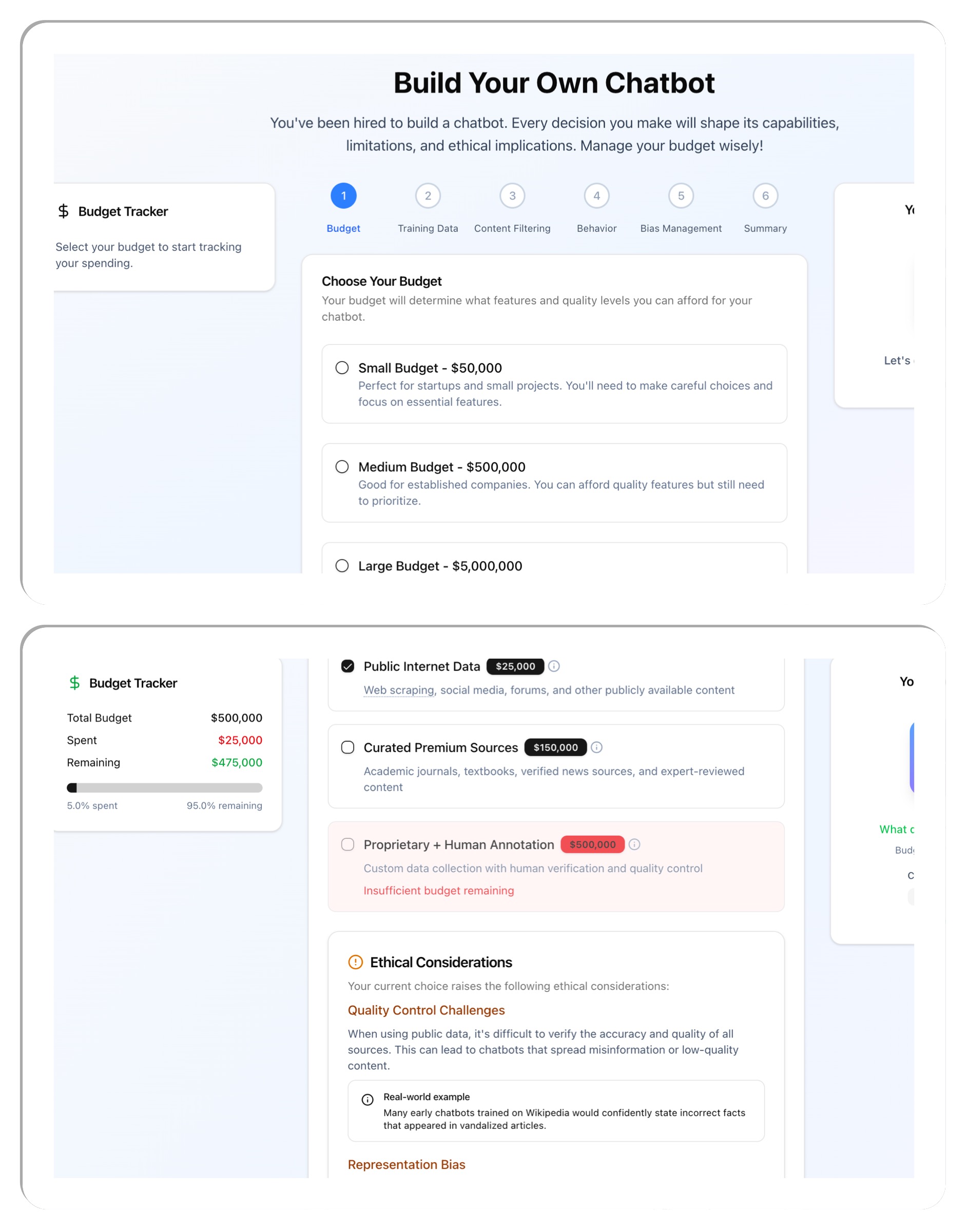}
\caption{Interactive chatbot, Groningen edition 2024/2025. Top: starting page with budget choice; bottom: data selection for training and warning associated with low-budget data choices. The budget gets progressively updated (see left). Credits: Marco Boasso, Kylian de Rooij, Emiel Dost, Andrew Geddes,  Stijn Schreven.\label{fig:chatbot}}
\end{figure}

\begin{figure}[htb]
\includegraphics[width=\linewidth]{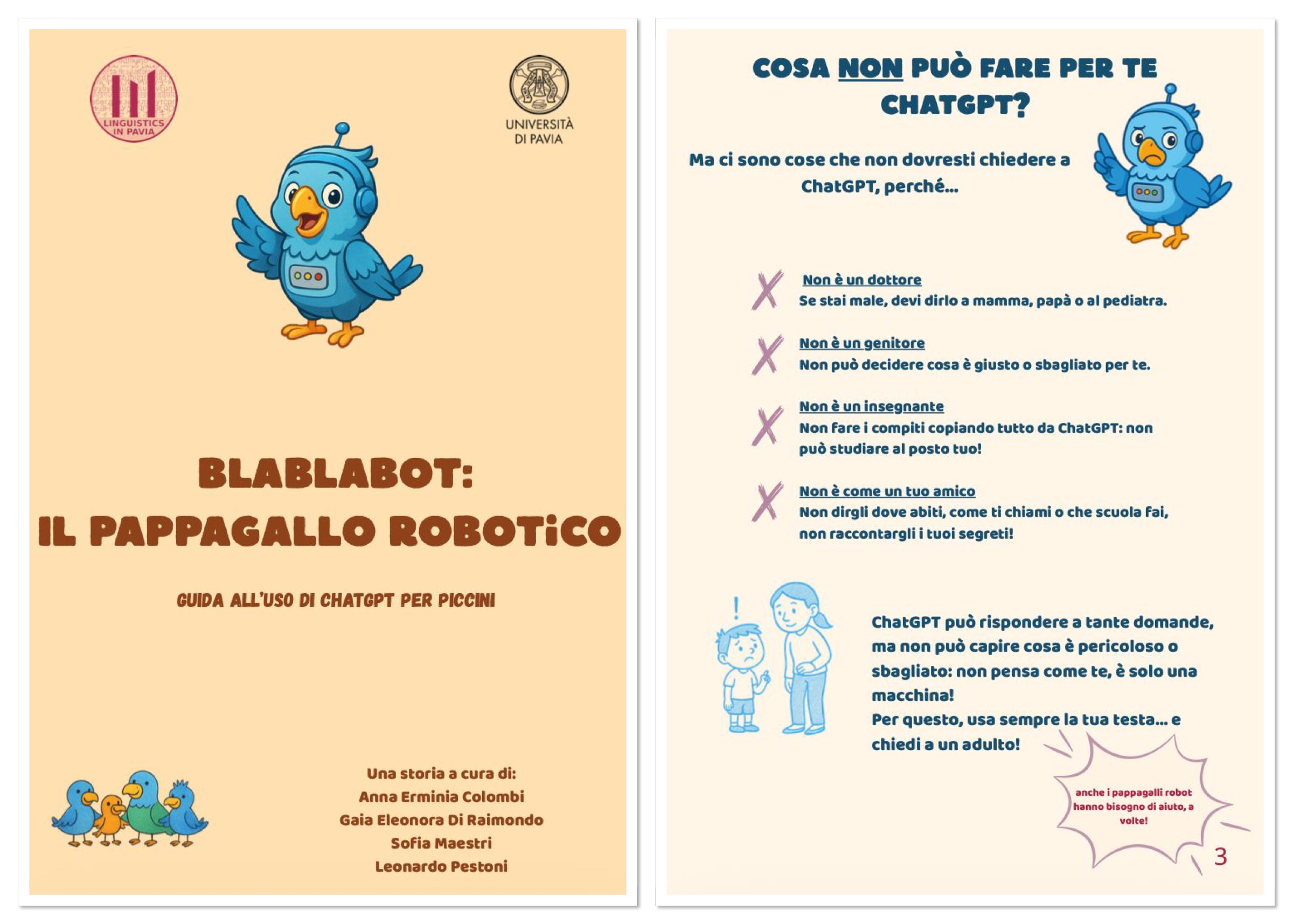}
\caption{Leaflet for school children, Pavia edition 2024/2025. On the left: cover page; on the right: one central page on how \textit{not} to use ChatGPT. Credits: Anna Erminia Colombi, Gaia Eleonora Di Raimondo, Sofia Maestri, Leonardo Pestoni.\label{fig:pavia-leaflet}}
\end{figure}

\begin{figure}[htb]
\includegraphics[width=\linewidth]{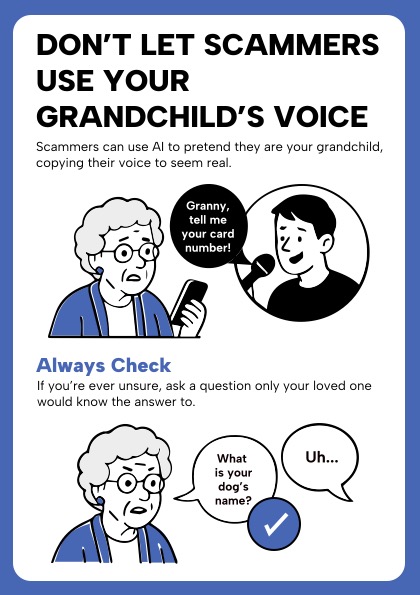}
\caption{Part of leaflet for elderly people in care homes, Turin edition 2024/2025. Follows explanation in other parts of the leaflet that talk about how state-of-the-art speech technology can now make use of one's voice in a very credible manner. Credits: Elina Saifutdinova, Evgeniya Voropaeva, Kseniia Zakharneva, Tatiana Semenova.\label{fig:leaflet-turin}}
\end{figure}

% \begin{figure}[htb]
% \includegraphics[width=\linewidth]{latex/interviews_turin.png}
% \caption{Screenshots from video of interviews with laypersons about the influence of AI in education,
% ecology and art, Turin edition 2024/2025. Credits: Chiara Falcioni, Martina Tazzini, Alex
% Tessarin, Emir Ünverdi.\label{fig:interviews-turin}}
% \end{figure}

%\subsection{Testimonials}

\begin{tcolorbox}[width=1\linewidth, colframe=gray, colback=blue!15!gray!15, boxsep=2mm, arc=3mm]
%\space{\Huge ``}
%\\
\textit{
``The course offered tools to move beyond polarisations through discussion in a space focused more on asking questions than giving answers, where the aim was not to take sides but to dig into the topic and probe it with a critical lens."}

%\hspace*{1.5em}{\Huge\hfill ''}
\smallskip

{\small A student in MA Linguistics (Pavia) [originally in Italian, translation is ours]}

\end{tcolorbox}

%\clearpage

\begin{figure*}[htb]
\centering
\fbox{\includegraphics[scale=.65]{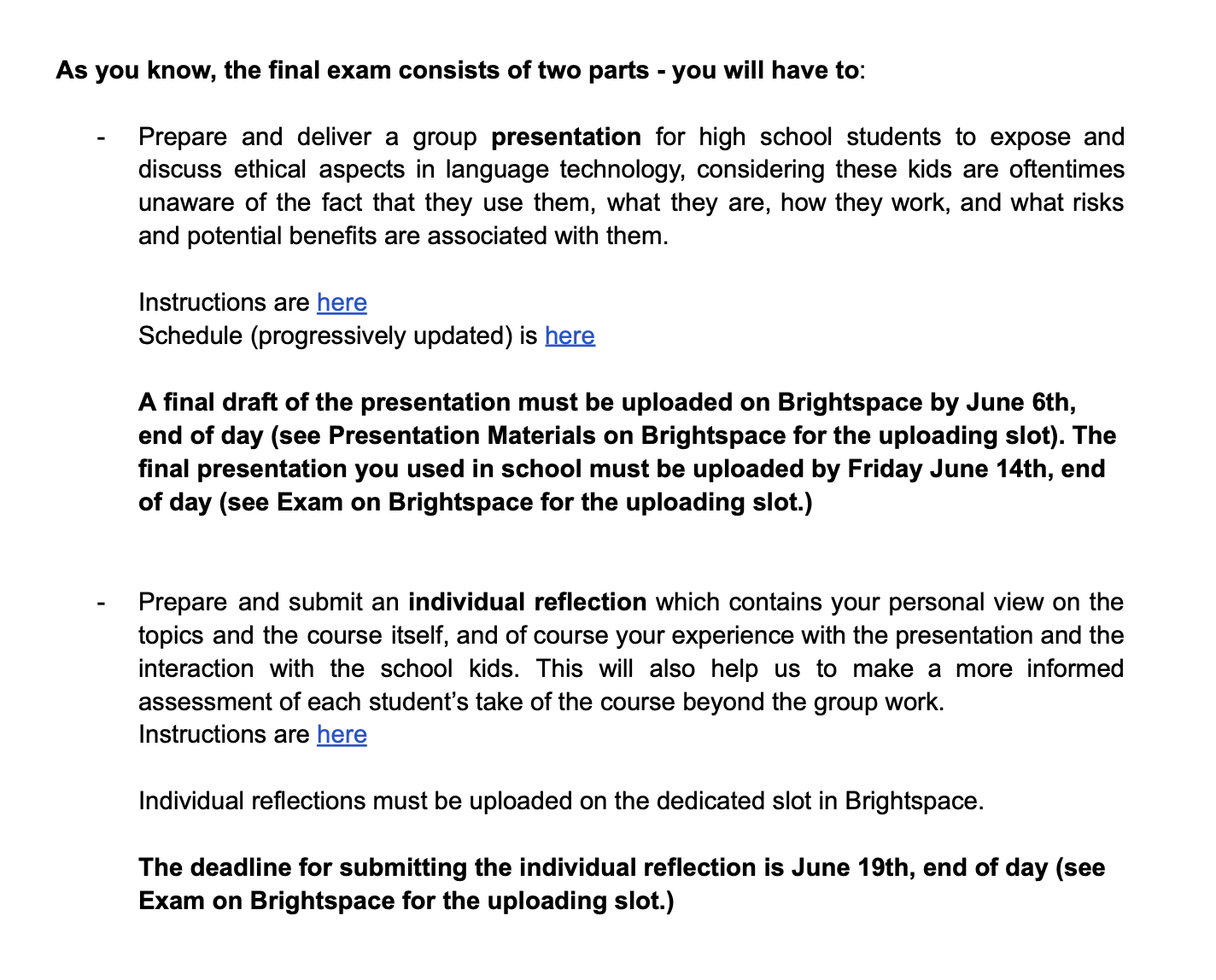}}
\caption{General exam instructions. Links point to details of presentations' and reflections' requirements and rubric (reported in Figures~\ref{fig:presentation} and \ref{fig:reflection} and schedule for school appointments per group with dates, times, and contacts.\label{fig:exam}}
\end{figure*}
%\includepdf[pages=-]{exam.pdf}

\begin{figure*}[htb]
\centering
\fbox{\includegraphics[scale=.5]{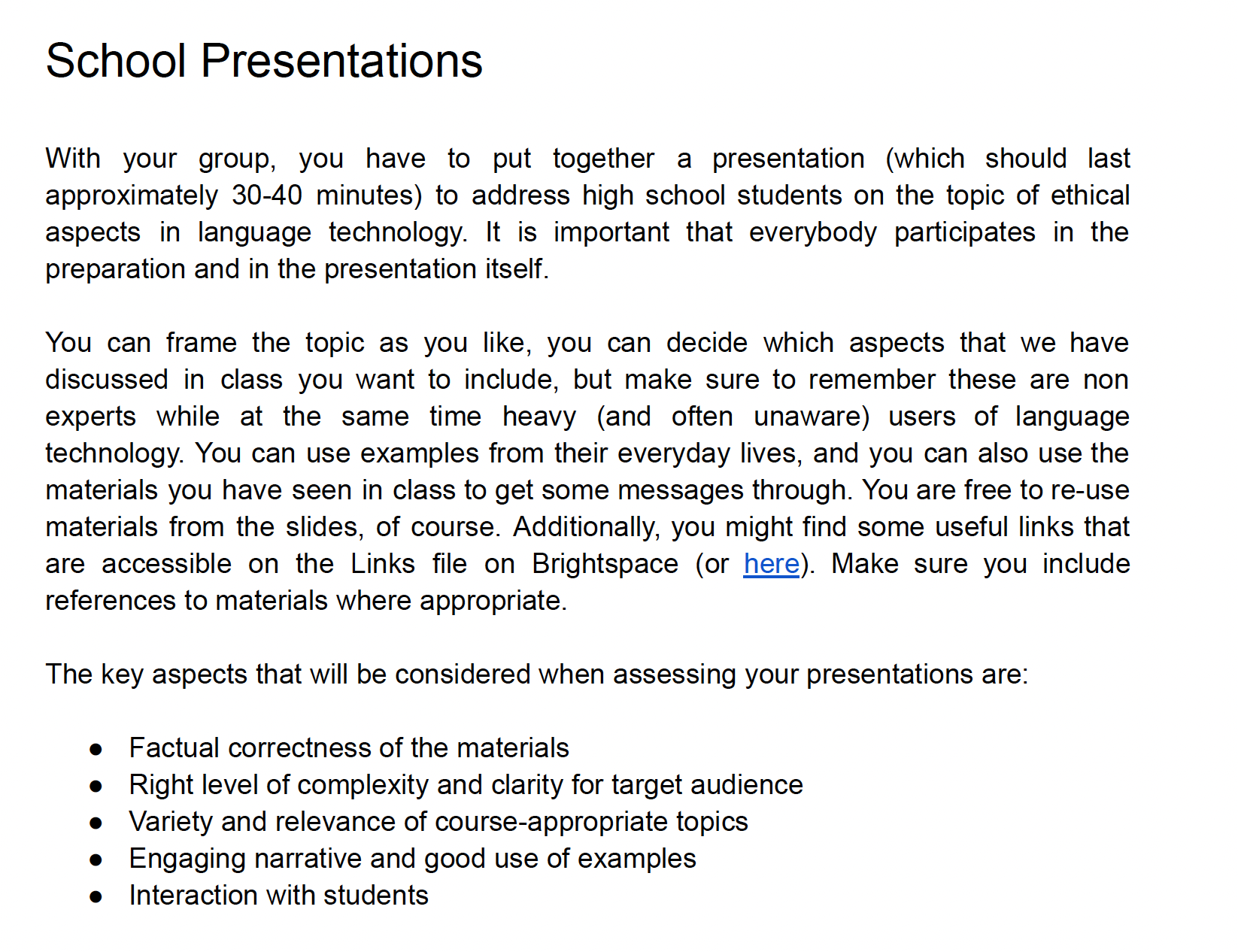}}
\caption{Presentation instructions.\label{fig:presentation}}
\end{figure*}

\begin{figure*}[htb]
\centering
\fbox{\includegraphics[scale=.7]{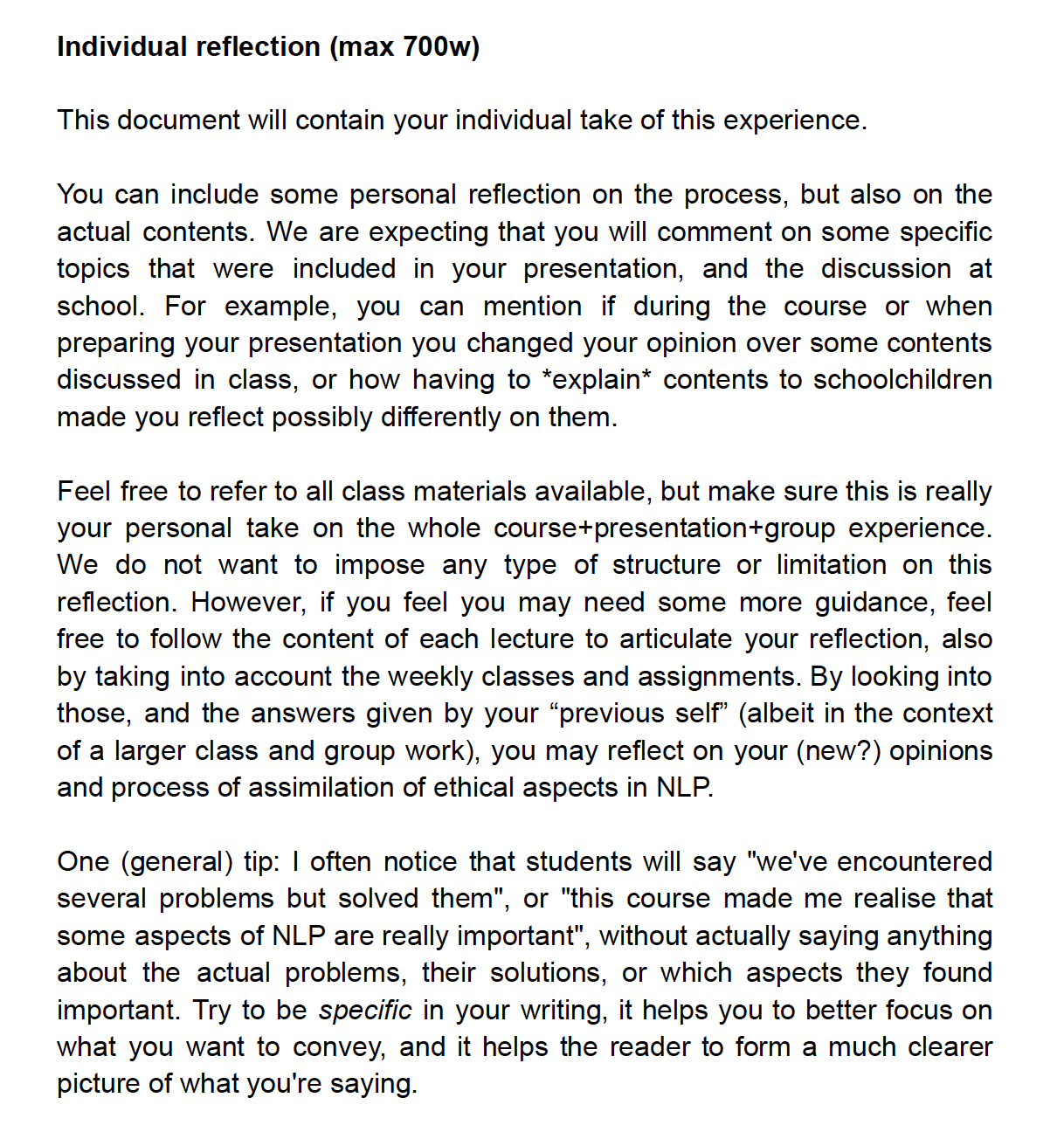}}
\caption{Individual reflection instructions.\label{fig:reflection}}
\end{figure*}

\clearpage



\section*{Supplementary material}

> **As you know, the final exam consists of two parts - you will have to**:
>
> - Prepare and deliver a group **presentation** for high school students to expose and discuss ethical aspects in language technology, considering these kids are oftentimes unaware of the fact that they use them, what they are, how they work, and what risks and potential benefits are associated with them.
>
>     Instructions are here
>     Schedule (progressively updated) is here
>
>     **A final draft of the presentation must be uploaded on Brightspace by June 6th, end of day (see Presentation Materials on Brightspace for the uploading slot). The final presentation you used in school must be uploaded by Friday June 14th, end of day (see Exam on Brightspace for the uploading slot.)**
>
> - Prepare and submit an **individual reflection** which contains your personal view on the topics and the course itself, and of course your experience with the presentation and the interaction with the school kids. This will also help us to make a more informed assessment of each student's take of the course beyond the group work.
>     Instructions are here
>
>     Individual reflections must be uploaded on the dedicated slot in Brightspace.
>
>     **The deadline for submitting the individual reflection is June 19th, end of day (see Exam on Brightspace for the uploading slot.)**

Figure 13: General exam instructions. Links point to details of presentations' and reflections' requirements and rubric (reported in Figures 14 and 15 and schedule for school appointments per group with dates, times, and contacts.

## School Presentations

With your group, you have to put together a presentation (which should last approximately 30-40 minutes) to address high school students on the topic of ethical aspects in language technology. It is important that everybody participates in the preparation and in the presentation itself.

You can frame the topic as you like, you can decide which aspects that we have discussed in class you want to include, but make sure to remember these are non experts while at the same time heavy (and often unaware) users of language technology. You can use examples from their everyday lives, and you can also use the materials you have seen in class to get some messages through. You are free to re-use materials from the slides, of course. Additionally, you might find some useful links that are accessible on the Links file on Brightspace (or [here](#)). Make sure you include references to materials where appropriate.

The key aspects that will be considered when assessing your presentations are:

- Factual correctness of the materials
- Right level of complexity and clarity for target audience
- Variety and relevance of course-appropriate topics
- Engaging narrative and good use of examples
- Interaction with students

Figure 14: Presentation instructions.

**Individual reflection (max 700w)**

This document will contain your individual take of this experience.

You can include some personal reflection on the process, but also on the actual contents. We are expecting that you will comment on some specific topics that were included in your presentation, and the discussion at school. For example, you can mention if during the course or when preparing your presentation you changed your opinion over some contents discussed in class, or how having to *explain* contents to schoolchildren made you reflect possibly differently on them.

Feel free to refer to all class materials available, but make sure this is really your personal take on the whole course+presentation+group experience. We do not want to impose any type of structure or limitation on this reflection. However, if you feel you may need some more guidance, feel free to follow the content of each lecture to articulate your reflection, also by taking into account the weekly classes and assignments. By looking into those, and the answers given by your "previous self" (albeit in the context of a larger class and group work), you may reflect on your (new?) opinions and process of assimilation of ethical aspects in NLP.

One (general) tip: I often notice that students will say "we've encountered several problems but solved them", or "this course made me realise that some aspects of NLP are really important", without actually saying anything about the actual problems, their solutions, or which aspects they found important. Try to be *specific* in your writing, it helps you to better focus on what you want to convey, and it helps the reader to form a much clearer picture of what you're saying.

Figure 15: Individual reflection instructions.

# Interview Template

The following are the topics we would like to ask you questions about.

- The ACL ethics committee
- Accountability, responsibility, and legal matters in NLP
- Some technical aspects
- Dominance of English in NLP
- Public policy and NLP

**ACL Ethics Committee (Ties)**

**1. On your website it says that much of your research group's work focuses on NLP for social good and that one of your goals is to enable NLP for diverse and disadvantaged users. Is this part of your motivation to be part of the ACL's Ethics Committee?**

*(self-explanatory)*

**2. Do you think that stricter ethics guidelines and awareness may lead to self-censorship and/or cause researchers to (unnecessarily) avoid certain research topics? It isn't far-fetched to imagine a scenario where some researchers may decide to avoid certain topics altogether so as to not attract any potential controversy or public outrage.**

*prof. dr. Nissim mentioned after the lecture that she fears that this may increasingly become an issue, and a pop article described a situation where a senior scientist talked to a young researcher who was doubting whether to continue their research.*

**3. Do you think there should be an ACL-wide, universal definition of bias because the ethics checklist explicitly leaves room for interpretation on what bias actually constitutes. Could a universal, structured, comprehensive guideline help avoid misalignment between researchers?**

*A significant part of the ethics checklist is devoted to bias, but explicitly leaves room for interpretation by paper submissions on what bias actually constitutes. Blodgett et al. (2020) discuss this topic, but do not suggest a concrete dictionary-like definition. A universal, structured, official, comprehensive guideline of the topic might help avoid misalignment between researchers.*

**4. Have you used the ethics guidelines and checklists in practice or have seen it being successfully used? If you did, then have you encountered issues in the practical application of the guidelines that should be addressed?**

*Throughout this week's course materials, the ethics guidelines popped up. However, we have seen very little concrete material about how it works in the field. We're curious as to how the interviewee experiences the new regulations.*

**Some technical aspects (Patrick)**

**1. Do you (and if so: how?) think corpora from non-digital origins (e.g. local books/newspapers/religious texts) can help de-bias language technologies?**

*As discussed in the first part of the assignment - how can we successfully increase the diversity of our training data? And will this help with de-biasing?*

**2. Do you know of any promising methods that may help separate between gender/race/class inappropriate (e.g. stereotypes) and appropriate (e.g. grammatical gender or contextual clues) language? Machine learning methods essentially learn from co-occurences of words. How do you separate legitimate from illegitimate co-occurences? Will this require linguistic knowledge, dictionary lookups, or perhaps modified training data to mitigate the symptoms?**

*(self-explanatory)*

**3. Do you think that de-biasing contextual clues in e.g. word embeddings and images could/should become part of a compulsory pre-processing pipeline? Papers on de-biasing frequently suggest de-biasing techniques which do not seem to affect performance too much. Should such practice become some kind of de-facto standard in NLP?**

*Two papers which we read are both concerned with the bias-reinforcing effect of contextual clues hidden in data. Both papers suggest de-biasing techniques which do not seem to affect performance too much. Should such practice become some kind of de-facto standard in NLP?*

**Accountability, responsibility, and legal matters in NLP (Patrick)**

**1. Do you think that public information (e.g. public posts from social media) should be in the public domain or should the author always retain the full ownfership rights to the text? Who do you think owns information generated through public discussions? What implications does this have legally and ethically?**

*This fundamental question was discussed during the lab as we talked about the ethics and legality of using public Tweets. If public space = public domain, then do people actually own their tweets and utterances? And if they do not: isn't the knowledge that everything you post publicly may be analyzed, scrutinized, and used for purposes that you wouldn't want highly uncomfortable? But then again, would this also apply for literature and opinion articles? And if not: how are those different from Tweets?*

**2. Most moral philosophers hold the view that the ability to reason and act independently is a requirement for accountability, and at this point in time this is not the case for AI systems. Who do you think is responsible for decisions made by NLP systems? Or: responsible for the consequences of NLP systems? Do you think the responsibility lies with the developers, the people who deploy them, or somewhere else?**

*A fundamental problem in ethics is that of moral agency: most philosophers hold that the ability to reason independently is a requirement for accountability. Currently, most AI systems do not really have that. We want to know who Julia thinks is responsible for their use.*


**Dominance of English in NLP (Jordy)**

**1. How would you feel about a quorum for the proportion of paper submissions that focus on an English corpus? Or: forcing papers to focus on a different language if the topic allows for it. Perhaps demanding that papers contain languages with various traits (e.g. non-Latin alphabet; complex morphology; flexible word order), would put pressure on researchers to actually deviate from using English-as-default.**

*Emily Bender and other researchers talk about the problem of the status of the English language as the "default" language in NLP. No concrete solution has been proposed, aside from explicitly naming the researched/used languages and having a public debate about the problem. A hard quotum of rule of some kind, perhaps not even regarding the use of English, but demanding that papers contain languages with various traits (e.g. non-Latin alphabet; complex morphology; flexible word order), would put pressure on researchers to actually make a change.*

**2. Do you (and if so: how?) think that multilingual data could be used to balance different cultural and linguistic biases? It's likely that Indian English, South African English, and American English all have different biases in their data.**

*Would the use of data other than American English data give an improvement or would it just introduce different problems due to the languages such as grammatical assumptions of gender. Could multilingual data be used to balance different language biases?*

**3. Do you think that the continuing status of English as the "default" language in NLP reinforces the Anglo-Saxon global cultural hegemony? Is this even a problem? Could multilingual data halt or reduce Anglo-Saxon dominance?**

*During the Machine Translation course we read a paper that explicitly warned for the risk of "global cultural hegemony" propagated through the widespread use of English. It begs the question whether English-first NLP research reinforces this problem.*

**Government and NLP (Ties)**

**1. Which do you think is the more pressing concern: biases in NLP, or the generation of fake reality (e.g. news, images, and voices)? Given limited time, would you focus on de-biasing or more controlled use of models?**

*These two aspects of AI (biases and the generation of fake reality) are subtly abused - much more subtly than other fields of AI such as (weaponized) robotics. Given limited resources, which would be the more pressing issue? Biases in NLP reinforce already existing biases, but uncontrolled use of generative tools may result in society beginning to live in different "realities of truth".*

**2. What do you think the role of the government should be in the deployment of NLP systems? Should there be legal boundaries in what is allowed and what not? What institution or regulatory body would enforce these laws?**

*<self explanatory>*

**3. How would you feel about your public research being incorporated in a multilingual robo-dog patrolling the streets of Shanghai or the US-Mexico border? What kind of responsibility do you feel regarding possible abuse of your publicly-funded work?**

*Public research will no doubt be incorporated into industrial applications. This week's pop articles talk about robo-dogs and military use. A researcher no doubt has feelings about this.*



% \begin{figure*}[htb]
% \centering
% \fbox{\includegraphics[scale=.65]{latex/interview1.png}}
% \caption{Sample interview template (each group made a different one, based on preferences and also target interviewee).\label{fig:interview}}
% \end{figure*}

\end{document}